\documentclass[letterpaper, 10 pt, conference]{ieeeconf}  

\IEEEoverridecommandlockouts                              

\usepackage{graphics} 
\usepackage{graphicx}
\usepackage{epsfig} 
\usepackage{mathptmx} 
\usepackage{times} 
\usepackage{amsmath} 
\usepackage{amssymb}  
\usepackage{bbm}
\usepackage{hyperref}

\newcommand{\norm}[1]{\left\lVert#1\right\rVert}

\title{\LARGE \bf
Baking in the Feature: Accelerating Volumetric Segmentation by Rendering Feature Maps 
}

\author{Kenneth Blomqvist$^{1}$, Lionel Ott$^{1}$, Jen Jen Chung$^{1, 2}$ and Roland Siegwart$^{1}$ 
\thanks{$^{1}$Autonomous Systems Lab, Swiss Federal Institute of Technology in Z\"urich, Switzerland. 
    {\tt\small kblomqvist@mavt.ethz.ch}}%
\thanks{$^{2}$School of ITEE, The University of Queensland, Australia.}%
}

\begin{document}

\maketitle
\thispagestyle{empty}
\pagestyle{empty}

\begin{abstract}
Methods have recently been proposed that densely segment 3D volumes into classes using only color images and expert supervision in the form of sparse semantically annotated pixels. While impressive, these methods still require a relatively large amount of supervision and segmenting an object can take several minutes in practice. Such systems typically only optimize their representation on the particular scene they are fitting, without leveraging any prior information from previously seen images.

In this paper, we propose to use features extracted with models trained on large existing datasets to improve segmentation performance. We bake this feature representation into a Neural Radiance Field (NeRF) by volumetrically rendering feature maps and supervising on features extracted from each input image. We show that by baking this representation into the NeRF, we make the subsequent classification task much easier.

Our experiments show that our method achieves higher segmentation accuracy with fewer semantic annotations than existing methods over a wide range of scenes.
\end{abstract}

\section{INTRODUCTION}

Neural Radiance Fields (NeRFs) \cite{mildenhall2020nerf} have recently emerged as a popular representation for 3D scenes due to their many favourable properties. They can accurately infer the geometry of a scene by making use of strong geometric structure and rich supervision from captured calibrated images. They have few hyperparameters to tune and are able to handle a wide range of scales, geometries and materials. 

As NeRFs use an MLP to map spatial coordinates to color values, they can easily be modified to predict other observable spatial properties. This led to NeRFs quickly being applied to semantic volumetric segmentation through SemanticNeRF \cite{zhi2021place}, which is able to propagate semantic pixel labels from one frame of a scene to another and across image pixels through generalization. This was subsequently demonstrated within an interactive semantic segmentation system, iLabel \cite{zhi2021ilabel}, motivating the use of such a system to generate ground-truth data for downstream embedded, real-time computer vision algorithms. 

While such systems achieve increasingly high accuracy as the amount of annotated pixels grows, they still require a lot of human supervision. SemanticNeRF used one labeled \emph{randomly selected} pixel per class, per image. With over 900 training images in a scene, this is a lot of annotated pixels from a diverse set of viewpoints. If used to annotate data, providing such supervision can take an infeasibly long amount of time. An inherent limitation of current semantic NeRF approaches is that they blindly map 3D scene coordinates to radiance, density, and semantic class by optimizing a randomly initialized neural network per scene. Structure priors or previously learned information is not leveraged. 

On the other hand, deep learning has fueled a whole body of work on representation learning, the goal often being to learn a feature representation that can be transferred to another task. Using pre-learned features in the context of NeRFs is not straightforward, as a NeRF maps scene-specific point coordinates to output values, which makes inducing desirable bias or structure into the model challenging. An example of this is \emph{object} bias. For example, if a user of a semantic annotation system clicks on a cookie in a kitchen scene, the user might ideally want the system to, by default, automatically segment the entire cookie from other cookies and the background, without having to select each pixel. 

\begin{figure}[t!]
\centering
\includegraphics[width=0.8\linewidth]{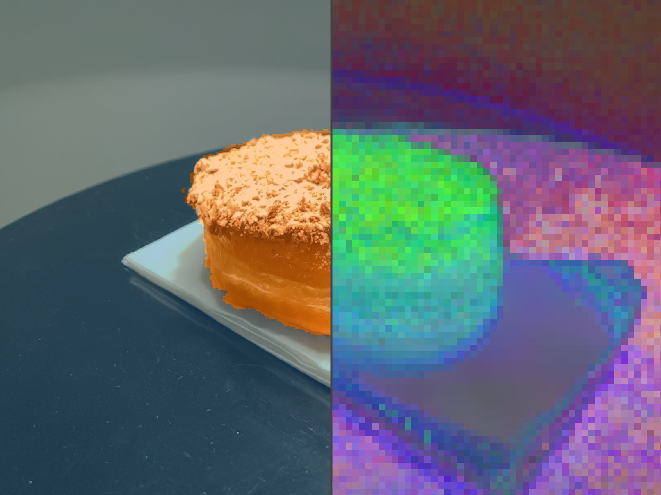}
\caption{The left side shows an example image from our dataset, overlayed with a segmentation mask produced by our system. The right side shows DINO features, which we reconstruct in our algorithm, mapped to RGB values using PCA.}
\label{fig:features}
\end{figure}

The goal of our work is to infer a dense 3D semantic segmentation of the scene, and infer dense 2D semantic segmentation maps for each input image, while spending as little expert annotation time as possible. To this end, we develop a user interface which allows a user to quickly segment RGB-D video sequences, by drawing sparse annotations on the images. While the user is using the system, the program infers a segmentation given the current user inputs and shows it to the user. The user can refine the segmentation, until they are happy with the result. 


To infer the best possible segmentation using a sparse set of labels, we propose a novel algorithm that models the scene using an implicit NeRF representation and leverages semantic image features obtained using a neural network feature extractor. We supervise our implicit scene model on these features, effectively \emph{baking} a feature representation into the learned MLP. We volumetrically render feature maps from the hidden layer activations of the semantic branch of the MLP, and minimize discrepancy between the rendered feature maps and extracted image feature vectors during training. This forces the MLP to encode additional structure, inducing desirable object, shape, and appearance bias into the learned representation, making the subsequent semantic classification much easier from sparse supervision. Our hypothesis is that such features encapsulate relevant spatial, object and semantic properties which are hard to learn by purely regressing color and radiance from position. Figure \ref{fig:features} illustrates how extracted features can encode information that can help distinguish between objects in a scene.

To summarize, our contributions are:
\begin{itemize}
\item A semantic NeRF algorithm that uses extracted image features to improve segmentation performance 
\item A hybrid feature encoding that is better suited for the semantic segmentation task
\item A volumetric segmentation system, including a graphical user interface, that can be used to quickly generate dense segmentation masks from sparse pixel annotations
\end{itemize}

In experiments we validate our pipeline on a diverse set of real-world scenes. We perform a larger scale evaluation on scenes from the Replica \cite{straub2019replica} dataset, which contain many more objects and semantic classes. On these datasets, we compare the performance against baseline methods, as well as using different feature extractors, namely Fully Convolutional Neural Networks \cite{long2015fully} and DINO \cite{dosovitskiy2020image}. Our results show that our DINO feature map supervised semantic NeRF formulation outperforms previously proposed semantic NeRFs on both accuracy and learning speed across all scenes, and can do so with much less human supervision. 

We make visual results, data and code available through the accompanying web page\footnote{\href{https://keke.dev/baking-in-the-feature/}{keke.dev/baking-in-the-feature}}.

\section{RELATED WORK}

\subsection{Automated Labeling}

Our goal is to infer a 3D segmentation of a scene and 2D semantic segmentation maps as quickly as possible. Several works have tackled a similar problem of inferring scene properties in an automated manner.

Object-based methods use knowledge of object geometry or category to speed up the workflow of annotating scenes. LabelFusion~\cite{marion2018label} introduced a tool to align an object model with a 3D scene with ICP refinement. While differentiable rendering was used to register the shape and pose of objects using shape priors in \cite{zakharov2020autolabeling} and \cite{beker2020monocular}. EasyLabel \cite{suchi2019easylabel} introduced a semi-automatic method for obtaining instance segmentations of 3D scenes by incrementally building up the scene. RapidPoseLabels \cite{singh2021rapid} presented a way to compute object pose and segmentation masks from sparse 2D keypoints, combining several pointclouds of an object. 

Other approaches have explored speeding up RGB-D data annotation by leveraging structure in the data. SALT \cite{stumpf2021salt} introduced a GrabCut \cite{rother2004grabcut} based approach to speed up labeling of RGB-D data. DeepExtremeCut \cite{maninis2018deep} computes dense object segmentation masks from extreme points of an object in an image.

\subsection{Neural Radiance Fields}

Mildenhall et al. \cite{mildenhall2020nerf} introduced NeRFs for novel view synthesis using volumetric rendering. Since then, NeRFs have been extended in various ways, DS-NeRF \cite{deng2022depth} incorporated depth measurements to speed up training, a modality which we also leverage.
SemanticNeRF \cite{zhi2021place} extended NeRF to also infer a semantic field from sparse pixel annotations and was extended with iLabel \cite{zhi2021ilabel}, a system for densely annotating scenes. We extend \cite{zhi2021place} to leverage representations which capture structure present in the input images to learn a better segmentation from fewer labels. While NeRF methods operate on a single scene, \cite{lazova2022control} introduced a NeRF model that learns a rendering module across many scenes. While they disregard semantic labels, their approach could be extended to tackle the segmentation problem. NeRF-Supervision \cite{yen2022nerf} used NeRFs to learn view-invariant dense object descriptors.

Other methods have extended semantic fields in other ways. PanopticNeRF \cite{fu2022panoptic} learns a semantic field from noisy 2D predictions and user-annotated 3D bounding boxes. Panoptic Neural Field \cite{kundu2022panoptic} learns several 4D neural fields that can reconstruct and track moving objects. NeSF \cite{vora2022nesf} similarly uses known calibrated images and NeRF to learn a semantic field of a scene. Instead of adding a semantic branch to the radiance field, they learn a 3D UNet to map density fields to semantic volumes. They learn the semantic field across many scenes, not targeting single scene label propagation. 

Concurrent work, N3F, \cite{tschernezkineural} uses neural rendering via \cite{tschernezki2021neuraldiff} and feature maps for one-shot object recognition, also showing some results for object segmentation. Similarly to them, we also use feature extractors and volumetric rendering to improve the learned 3D feature representation.
However, we use an autoencoder to match the feature dimensionality of the MLP model with the dimensionality of the features whereas they modify their architecture to render features with the same dimensionality as the input features. \cite{tschernezkineural} focuses on video understanding, while we are uniquely interested in speeding up human-in-the-loop segmentation to generate training data for downstream robotic vision tasks. Furthermore, \cite{tschernezkineural} uses a thresholding approach to object segmentation as opposed to propagating sparse pixel labels by inferring a categorical distribution over classes.

\section{METHOD}

\begin{figure}[t]
   \includegraphics[width=\linewidth]{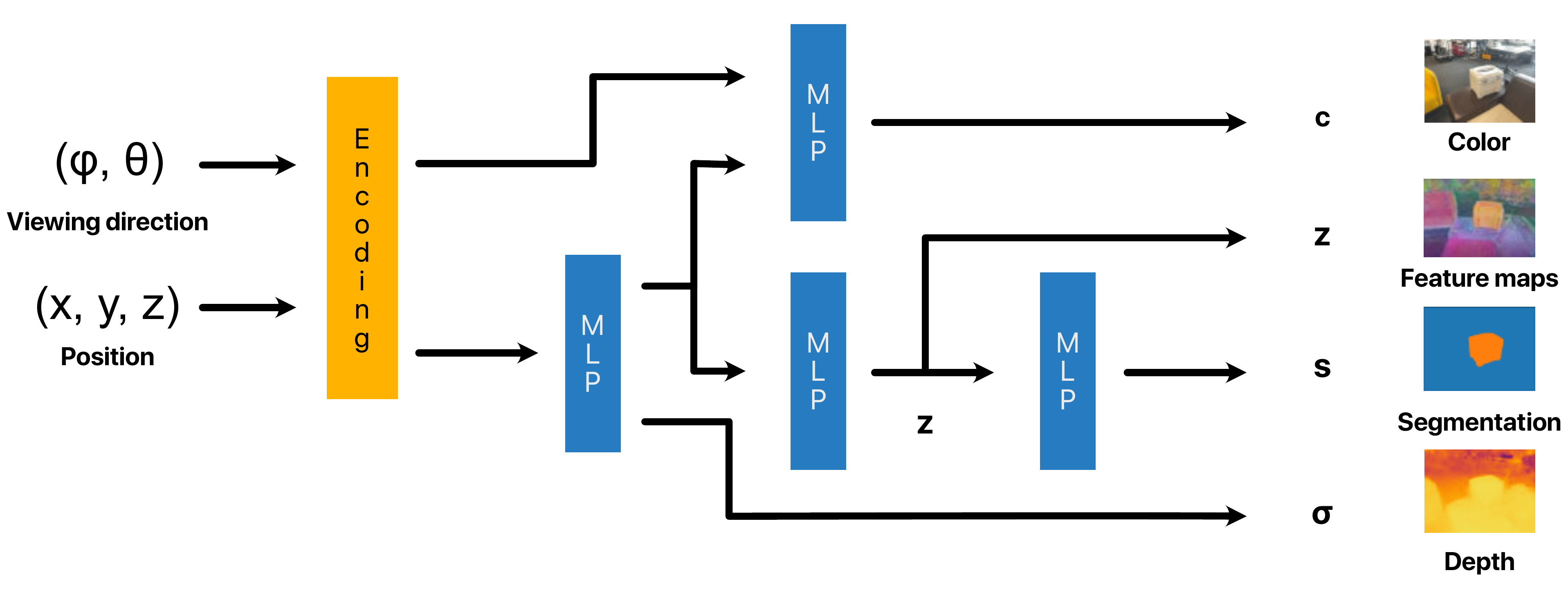} 
   \caption{Architectural diagram of our model. Scene coordinates and viewing direction are mapped to RGB, density, depth, feature maps and segmentation maps.}
   \label{fig:architecture}
\end{figure}

We design a NeRF-style algorithm, that maps each point in the scene to color, density, depth, semantic class, and semantic feature vector. Figure \ref{fig:architecture} shows a high-level diagram of our model, which consists of five main components: a feature encoder, a geometry MLP, a color MLP, a feature MLP, and a semantic classifier MLP. In this section, we first describe how we encode scene positions into feature vectors that are fed to the neural network. We then describe how use volumetric rendering to compute 2D image maps from the 3D representation and describe how we learn the parameters of the scene representation from provided data. Finally, we describe our graphical user interface, which allows a user to interact with the system.  

\begin{figure*}[t]
\includegraphics[width=\linewidth]{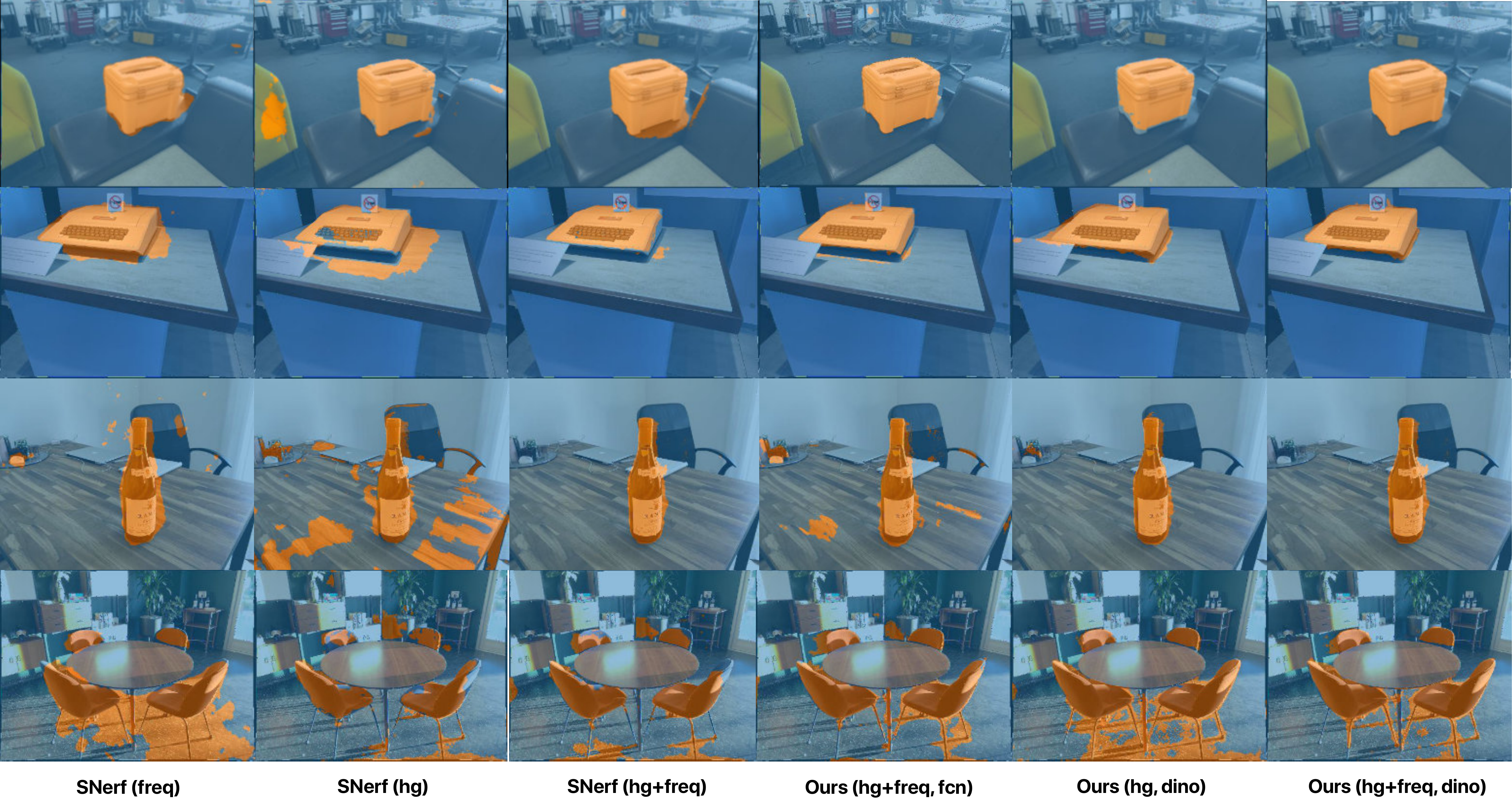}
\caption{Segmentation masks produced on the box scene using the different methods with the same set of annotations. The frequency encoding allows for spatial context while the hash grid encoding allows for a high level of detail locally. The feature supervision again helps in reasoning about object boundaries.}
\label{fig:masks}
\end{figure*}

\subsection{Positional Encoding}

NeRF \cite{mildenhall2020nerf} uses a frequency encoding based on sine and cosine functions:
\begin{equation}
\gamma(y) = [\sin{2^0\pi y}, \cos{2^0\pi y}, \dots, \sin{2^{L-1} \pi y}, \cos{2^{L-1} \pi y}],
\end{equation}
where $y$ corresponds to any of the 3D scene coordinates or a viewing direction which are normalized to the range $[-1, 1]$. $L$ defines the number of frequencies used. We use $L=10$ for 3D coordinates and $L=4$ for the viewing directions.

An alternative hash grid encoding was introduced in \cite{muller2022instant}, which greatly speeds up the learning of NeRFs. They define a hierarchical voxel grid over the scene coordinates, with parameters at cell vertices for every level. To encode a point, parameters are looked up at each level of the grid, trilinearly interpolated and concatenated to produce an encoding. These parameters are learned jointly with those of the MLP radiance field. 

While the hash grid encoding produces great visual results for view synthesis and greatly speeds up training and rendering, it is not optimal for our application, as simply using the hash grid encoding causes the model to overfit to the sparse user annotations. The resulting segmentation maps fail to infer the spatial relation between annotated points, and large areas not belonging to objects are assigned the wrong semantic class. See Figure \ref{fig:masks} for an illustration.

To solve this problem, we propose a hybrid approach using both the hash grid encoding concatenated with low frequency positional encoding with $L=2$. This enables the model to both reason about coarse spatial location, while the grid parameters allow for encoding finer details in the scene. As we are targeting an interactive system, using only frequency encoding would not be an option, as it requires many more iterations to learn high frequency details. 


\subsection{Neural Radiance Fields}

To integrate the scalar and vector outputs of our spatial field, we define a function $R$ to volumetrically render vector or scalar values from a function $h$ along a given ray $\mathbf{r}$:
\begin{align}
\label{eq:render}
R(\textbf{r}, h) &= \sum_{i=1}^N T_i (1 - \exp(-\sigma_i \delta_i)) h(\textbf{x}_i),\\
T_i &= \exp(-\sum_{j=1}^{i-1} \sigma_j \delta_j),
\end{align}
where $T_i$ is the transmittance function measuring the amount light transmitted through the ray $\mathbf r$ to sample $i$, $\delta_j$ is the distance between samples and $\sigma_i$ is the predicted density for encoded point samples $\mathbf{x}_i$ along the ray $\mathbf{r}$.  
We use this rendering function to render pixel color values, depth, semantic outputs and image features for a given ray $\mathbf{r}$:
\begin{align}
\hat{\mathbf{c}}(\mathbf{r}) &= R(\mathbf{r}, \mathbf{c}), \\
\hat{d}(\mathbf{r}) &= R(\mathbf{r}, d), \\
\hat{\mathbf{s}}(\mathbf{r}) &= R(\mathbf{r}, \mathbf{s}), \\
\hat{\mathbf{f}}(\mathbf{r}) &= R(\mathbf{r}, \mathbf{f}),
\end{align}
using the following functions: $\mathbf{c}$ for color, $d$ for depth, $\mathbf{s}$ for the semantic vector, and $\mathbf{f}$ for the intermediate feature vector of our model for encoded point samples along a ray $\mathbf{r}$.

We define the same photometric loss $L_{rgb}$ as in NeRF \cite{mildenhall2020nerf} and a depth loss $L_d$ similar to the one used by \cite{deng2022depth}:
\begin{align}
L_{rgb}(\mathbf{r}) &= \norm{\hat{\mathbf{c}}(\mathbf{r}) - \bar{\mathbf{c}}(\mathbf{r})}^2_2, \\
L_{d}(\mathbf{r}) &= \begin{cases}
\norm{{\hat{d}(\mathbf{r})} - \bar{d}(\mathbf{r})}_1, & \text{if $\bar{d}$ is defined for $\mathbf{r}$} \\
0, & \text{otherwise}
\end{cases}
\end{align}
where $\bar{\mathbf{c}}(\mathbf{r})$ is the ground truth and $\hat{\mathbf{c}}(\mathbf{r})$ the predicted color for ray $\textbf{r}$, $\bar{d}(\mathbf{r})$ is the ground truth depth (if available) and $\hat{d}(\mathbf{r})$ the integrated depth predictions along ray $\mathbf{r}$.

To learn the semantic class, we define a cross entropy loss:
\begin{equation}
L_{s}(\mathbf{r}) = \begin{cases}
-\log{ \frac{\exp(\hat{s}(\mathbf{r})_{\bar{s}(\mathbf{r})})}{\sum_{c=1}^{C_n} \exp(\hat{s}(\mathbf{r}))_c) } }, & \text{if $\bar{s}$ is defined for $\mathbf{r}$} \\
0, & \text{otherwise} \\
\end{cases}
\end{equation}
where $\bar{s}(\mathbf{r})$ is the ground truth class for ray $\mathbf{r}$, $\hat{\mathbf{s}}(\mathbf{r})$ is the integrated semantic MLP outputs and $\hat{s}(\mathbf{r})_{c}$ the output corresponding to class $c$.

\subsection{Learning Feature Maps}

We assume that we have a feature extractor, which maps images $\mathbf{I}^{H_i \times W_i \times 3}$ to feature maps $\bar{F}^{H_f \times W_f \times M}$ containing feature vectors $\bar{\mathbf{f}}$. $\mathbf{I}$ is an input image with height $H_i$ and width $W_i$, $\bar{F}$ has height $H_f$ and width $W_f$ and $M$ is the dimensionality of the feature vectors. The purpose of the feature extractor is to encode semantic and spatial information about a particular view, providing contextual information that cannot be inferred from a single scene, but can be learned by observing other data. Such functions include Vision Transformers \cite{dosovitskiy2020image} or Fully Convolutional Neural Networks \cite{long2015fully} that are learned on large datasets. 

To distill information in the feature maps into our 3D scene representation and to inform a downstream classifier, we propose to volumetrically render feature outputs $\mathbf{f}$ along a ray using \eqref{eq:render} to produce rendered feature maps $\hat{\mathbf{f}}$ and supervise on corresponding image features $\bar{\mathbf{f}}$. 


As the dimensionality of the image features $\bar{\mathbf{f}}$ may not match that of the rendered features $\hat{\mathbf{f}}$, and to reduce memory use and the amount of floating point operations when using our system, we use a simple MLP autoencoder to reduce the dimensionality of the image features $\bar{\mathbf{f}}$. The autoencoder has an encoder $enc$ and decoder $dec$. The encoder maps feature vectors with $M$ dimensions to $D$ dimensions and the decoder maps them back to $M$ dimensions. We set $D$ to 64 throughout our experiments. We fit the autoencoder by minimizing a standard reconstruction loss:
\begin{equation*}
   L_{ae}(\bar{F}) = \norm{dec(enc(\bar{F}(\textbf{p}))) - \bar{F}(\textbf{p})}_2 + \lambda_{ae} \norm{enc(\bar{F}(\textbf{p})}_1,
\end{equation*}
where $\bar{F}(\textbf{p})$ is the feature corresponding to sampled pixel $\textbf{p}$, and $\lambda_{ae}$ is a weight for the sparsity term. We use an $L_2$ loss to minimize information loss. The sparsity term is to encourage a sparse feature representation that can easily be classified into different classes.

As image features only depend on raw input images and a given pre-trained feature extractor, we pre-compute the autoencoder offline before the user interacts with the system and keep them fixed while the scene representation is learned.

When fitting a scene in our volumetric segmentation pipeline, to bake the feature representation $\hat{\mathbf{f}}$ into the scene representation, we define an additional feature loss:
\begin{equation}
L_{f}(\mathbf{r}) = \norm{enc(\bar{\mathbf{f}}(\mathbf{r})) - \hat{\mathbf{f}}(\mathbf{r})}_1,
\end{equation}
where $\mathbf{r}$ is an image ray, $\bar{\mathbf{f}}(\mathbf{r})$ is the image feature corresponding to ray $\mathbf{r}$, $\hat{\mathbf{f}}(\mathbf{r})$ is the rendered feature for ray $\mathbf{r}$. Should $H_i$ and $H_f$ differ, during training, we use nearest neighbor interpolation to lookup the corresponding image feature.

\subsection{Optimization and Sampling}

We optimize the combined loss using stochastic gradient descent using Adam over rays $\mathbf{r}$ sampled from the images:
\begin{equation}
L(\mathbf{r}) = L_{rgb}(\mathbf{r}) + \lambda_{d} L_d(\mathbf{r}) + \lambda_s L_{s}(\mathbf{r}) + \lambda_f L_{f}(\mathbf{r}),
\end{equation}
to jointly fit the positional encoding and MLP parameters.

As most pixels do not have a semantic class associated with them, we use a sampling scheme to balance the task of predicting a semantic class with the other objectives. When sampling examples for optimization, we select half the samples uniformly from all pixels. For the remaining half, to not bias the resulting function towards any class, we first select a class uniformly from the available classes. A pixel is then sampled from all pixels which are annotated with the sampled class. 

\subsection{Graphical User Interface}

\begin{figure}
\centering
\includegraphics[width=0.9\linewidth]{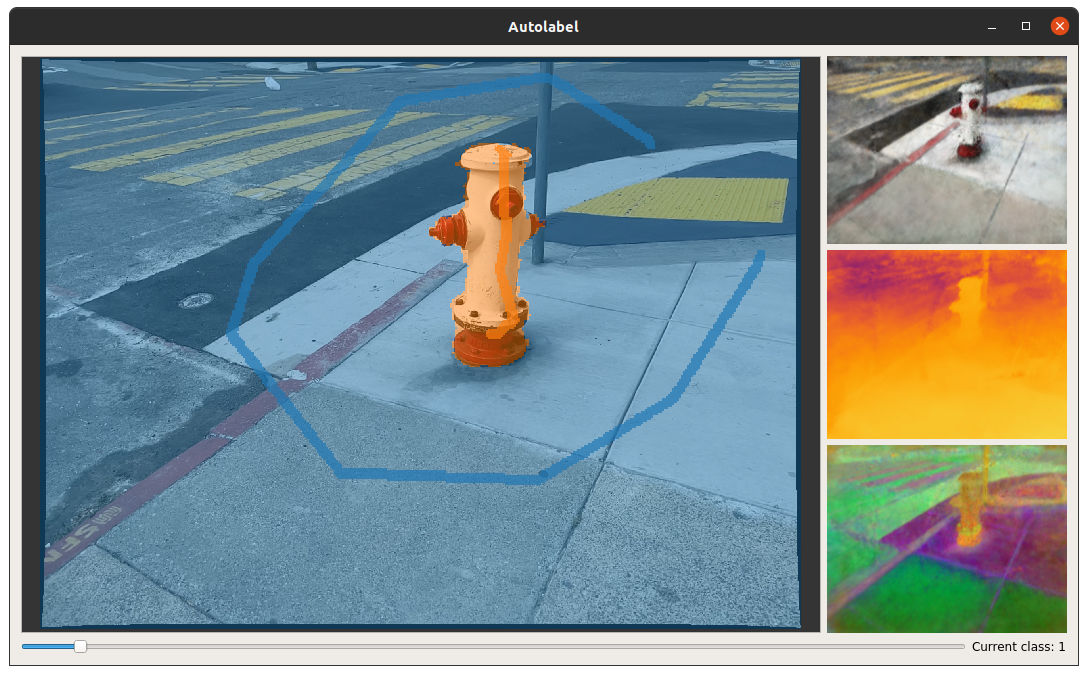}
\caption{The graphical user interface of our system. The user gives sparse annotations by drawing on images in a scan with the appropriate class. The system infers a volumetric segmentation given the current sparse annotations from all views and renders a dense segmentation map which the user can correct with further annotations.}
\label{fig:gui}
\end{figure}

Figure \ref{fig:gui} shows the user interface of our system. The user can move through frames in a captured RGB-D video sequence and draw annotations, shown as opaque lines, while the system fits a model to the scene and annotations. The translucent mask shows the current segmentation, which the user can correct. The right side shows rendered color, depth and features mapped to RGB using PCA.

\section{EXPERIMENTS}

\begin{table*}[t]
\small
\centering
\begin{tabular}{l c c c c c c}
Scene & SNeRF (freq) & SNeRF (hg) & SNerf (hg+freq) & Ours (hg+freq, fcn) & Ours (hg, dino) & Ours (hg+freq, dino) \\
\hline
apple 2             & 0.744 & 0.565 & 0.853 & 0.903 & 0.878 & \textbf{0.942} \\
bench               & 0.803 & 0.817 & 0.816 & 0.818 & 0.873 & \textbf{0.912} \\
box                 & 0.901 & 0.686 & 0.868 & 0.951 & 0.922 & \textbf{0.962} \\
chairs              & 0.531 & 0.635 & 0.705 & 0.686 & 0.696 & \textbf{0.761} \\
cup                 & 0.837 & 0.514 & 0.586 & 0.565 & 0.854 & \textbf{0.934} \\
doughnut            & 0.653 & 0.576 & 0.502 & 0.673 & 0.879 & \textbf{0.910} \\
fire hydrant 1      & 0.838 & 0.561 & 0.792 & 0.805 & 0.751 & \textbf{0.887} \\
fire hydrant 2      & 0.757 & 0.216 & 0.853 & 0.598 & 0.848 & \textbf{0.890} \\
hat                 & 0.911 & 0.937 & \textbf{0.960} & 0.950 & 0.919 & 0.959 \\
keyboard            & 0.895 & 0.699 & 0.900 & 0.926 & 0.943 & \textbf{0.947} \\
shoe                & 0.791 & 0.814 & 0.833 & 0.894 & 0.970 & \textbf{0.979} \\
valve               & 0.588 & 0.250 & 0.691 & 0.721 & 0.692 & \textbf{0.730} \\
wine bottle red     & 0.523 & 0.362 & 0.521 & 0.794 & 0.690 & \textbf{0.856} \\ 
wine bottle white   & 0.569 & 0.233 & 0.323 & 0.573 & 0.830 & \textbf{0.884} \\
\hline
Average & 0.739 & 0.560 & 0.732 & 0.778 & 0.703 & \textbf{0.897} \\
\end{tabular}
\caption{Intersection-over-Union agreement of produced segmentation maps on our captured scenes against manually annotated frames.}
\label{table:scenes}
\end{table*}

In our experiments we investigate and answer the following questions:
\begin{itemize}
    \item Does our method improve accuracy of produced segmentations given the same amount of supervision?
    \item Does supervising on features reduce the amount of annotations required to produce accurate dense segmentation maps for a scene?
    \item Does our algorithm improve labeling efficiency when used as part of an interactive labeling system? 
\end{itemize}

\subsection{Baselines}

We compare our algorithm against a baseline SemanticNeRF model \cite{zhi2021place}. We also compare the effect of using different positional encodings with both our algorithm and the baseline. For the positional encoding, \textbf{freq} refers to the frequency positional encoding in which case we use $L=10$ frequencies for the position, \textbf{hg} refers to the hash grid encoding and \textbf{hg+freq} is hash grid encoding concatenated with low frequency positional encoding ($L=2$).

As our algorithm can make use of any image features, we experiment with features from a Fully Convolutional Network \cite{long2015fully} trained on COCO \cite{lin2014microsoft} on the semantic segmentation task, denoted \textbf{fcn}. We compare this to features extracted using the DINO ViT-S/8 vision transformer model \cite{caron2021emerging} trained on ImageNet, denoted \textbf{dino}. 

To make the comparison fair, all baselines and our method use the same sampling and training pipelines. They simply differ in the scene model and loss functions used.

\subsection{Label Propagation Quality}


To investigate the first question, we scan a number of scenes using an RGB-D camera and run them through \cite{schonberger2016structure} to obtain camera poses. We annotate the scenes using the GUI by drawing squiggles on pixels belonging to each desired semantic class on 2 to 10 individual images, depending on the scene. What the annotations look like from a single view can be seen in Figure \ref{fig:gui}. From these annotations, the algorithms are tasked to segment the scene and infer what the user considers belonging to each class. All algorithms receive the exact same set of annotations. We run each algorithm on each scene and produce semantic segmentation maps for all images. We compare the produced segmentation maps against ground truth masks, obtained by labeling a reference subset of the images with a polygon mask for each object. We then compute the Intersection-over-Union agreement between the inferred and reference masks. 

It should be noted that results on this experiment are not indicative of the performance at the limit on dense segmentation maps, rather they measure the ability of the algorithms to generalize and figure out where object boundaries lie from a specific set of reasonable, sparse annotations.

\subsection{Human-in-the-loop Simulation}

\begin{figure*}[t]
\includegraphics[width=\linewidth]{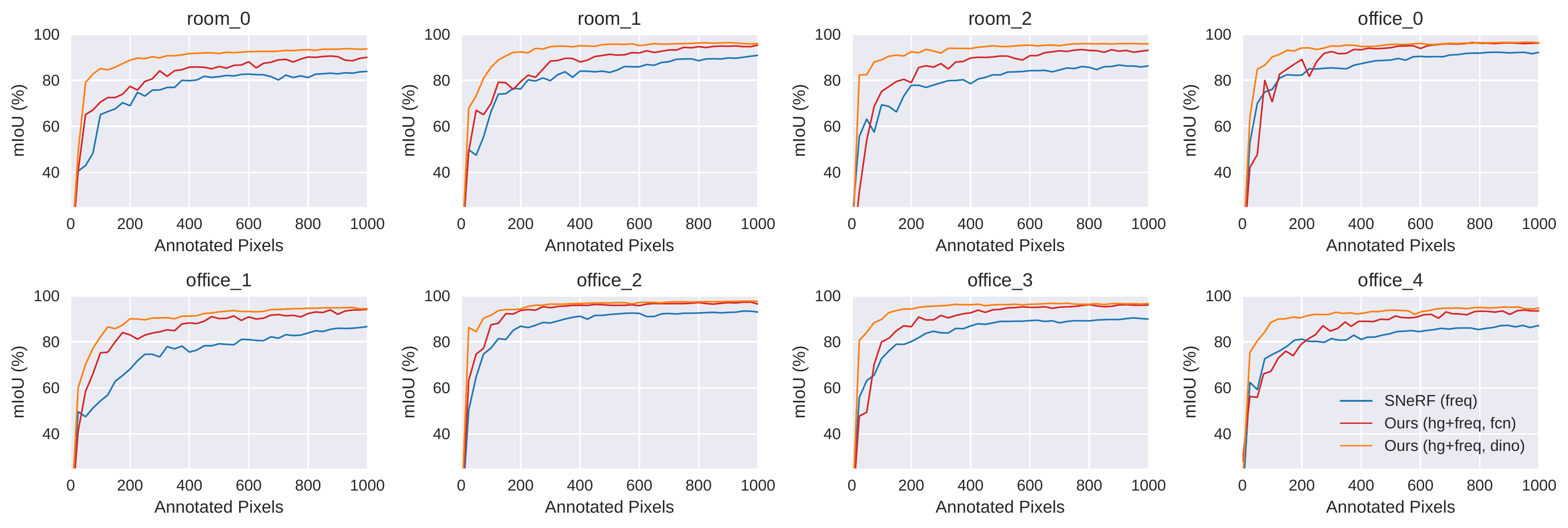}
\caption{Performance on different scenes of the Replica dataset. Our method outperforms baseline methods on all scenes for both learning speed and final accuracy.}
\label{fig:replica} 
\end{figure*}

As our system is designed to quickly create annotated data for downstream learning algorithms with little human input, we design a simulated human-in-the-loop experiment to test the algorithms in a fair, scalable, and reproducible way. As a dataset, we use the openly available Replica dataset \cite{straub2019replica} and more specifically, the rendered sequences published by \cite{zhi2021place}, which include dense ground truth semantic annotations. 

We initialize each algorithm by training only on RGB, depth and features for 15k iterations. We then run an iterative process predicting semantic labels using the current model parameters and select five pixels for which semantic classes are falsely inferred and add their ground truth labels to the training set. We then run 250 optimization steps with all labels accumulated thus far, as would be possible between user actions, and repeat the process.

We record the intersection-over-union agreement with the ground truth segmentation masks at each iteration step to observe how quickly the label propagation algorithm is able to produce high quality semantic labels. The ideal algorithm would converge to perfect labels after one click per object class by the user.

\section{RESULTS}
Table \ref{table:scenes} shows IoU agreement between manually labeled individual frames and the segmentations produced by the different methods. The best accuracy is achieved using hybrid encoding, supervised by DINO features on virtually all scenes. 

As illustrated in Figure \ref{fig:masks} and quantified in Table \ref{table:scenes}, using only hash grid encoding causes the model to overfit to the sparse annotations. Large areas outside of objects are assigned to the object class. This is especially apparent on the white wine bottle scene (Figure \ref{fig:masks}, image row 3, column 2). Using hash grid encoding with DINO feature supervision removes some of the errors, as the model is able to reason about visual and contextual properties of different areas in the scene. However, object boundaries are not captured as sharply when not using feature supervision. The DINO features perform much better than FCN features, indicating that the choice of feature extractor, is an important consideration.

\subsection{Human-in-the-loop}

Figure \ref{fig:replica} shows results on different scenes of the Replica dataset for the human-in-the-loop simulation. Our method performs better at the limit than the baseline methods on all of the scenes. The results show that the biggest benefit of feature supervision is that the model learns from much fewer labels, yet still achieves higher accuracy as more pixels are annotated.

The baseline SNeRF (freq) method takes on average over 5x longer to reach 80\% IoU accuracy compared to our method with DINO features and hybrid positional encoding. This means the user would have to spend less than a fifth of the clicks while using the system to achieve the same level accuracy. 

\section{CONCLUSIONS}

We presented an algorithm for volumetric segmentation, which we showcased in a human-in-the-loop data annotation and label propagation application. Our algorithm significantly speeds up learning and improves performance over baseline methods across our experiments. We performed an ablation study showing how different parts of our algorithm contribute to the overall performance. We demonstrated how using only hash grid encoding can cause a segmentation model to overfit to sparse user annotations and showed how this problem can be overcome by combining frequency encoding with hash grid encoding. 

As shown by our experiments, the choice of feature is important. Learning a better feature representation that is viewpoint invariant, yet allows for efficient segmentation of objects presents a promising research direction. Furthermore, our system assumes a static scene. Removing this limitation, is left as future work.

Our system produces high quality segmentation maps, but in many robotic applications computing other properties, such as object pose \cite{marion2018label}, shape \cite{gao2021kpam} or keypoints \cite{blomqvist2022semi} might be required. A framework which could with little input produce high quality segmentations in combination with object information, would be a breakthrough to fuel the learning-based perception algorithms proposed in recent years. 



\section*{ACKNOWLEDGMENT}
This project has received funding from EU Horizon 2020 program, project PILOTING H2020-ICT-2019-2 871542.

\bibliography{references}
\bibliographystyle{ieeetr}

\end{document}